\documentclass{article}

\usepackage{times}
\usepackage{graphicx} % more modern
\usepackage{subfigure} 

\usepackage{url}
\usepackage{latexsym}

\usepackage{amsmath}
\usepackage{amssymb}

\usepackage{natbib}

\usepackage{algorithm}
\usepackage{algorithmic}

\renewcommand\algorithmicthen{} % removes then and do from then and for algorithm
\renewcommand\algorithmicdo{}

\newcommand{\algrule}[1][.2pt]{\par\vskip.5\baselineskip\hrule height #1\par\vskip.5\baselineskip}

\def\bs#1{\boldsymbol{#1}}

\usepackage{hyperref}

\usepackage[accepted]{icml2014}

\begin{document} 

\twocolumn[
\icmltitle{Learning Semantic Script Knowledge with Event Embeddings}

% It is OKAY to include author information, even for blind
% submissions: the style file will automatically remove it for you
% unless you've provided the [accepted] option to the icml2014
% package.
\icmlauthor{Ashutosh Modi}{ashutosh@coli.uni-sb.de}
\icmladdress{Saarland University, Saarbr{\"u}cken, Germany}

\icmlauthor{Ivan Titov}{titov@uva.nl}
\icmladdress{University of Amsterdam, Amsterdam, the Netherlands}

\icmlkeywords{event embeddings, distributed representations}

\vskip 0.3in
]

\begin{abstract} 
Induction of common sense knowledge about prototypical sequences of events has recently received much attention~(e.g., \cite{chambers2008unsupervised,regneri2010learning}).
Instead of inducing this knowledge in the form of graphs, as in much of the previous work, in our method,   distributed representations of  event realizations are computed based on distributed representations of predicates and their arguments, and then these representations are used to predict prototypical event orderings.
The parameters of the compositional process for computing the event representations and the ranking component of the model are jointly estimated from texts.  
We show that this approach
results in a substantial boost in ordering performance with respect to  previous methods.
\end{abstract} 

\section{Introduction}

It is generally believed that natural language understanding systems would benefit
from incorporating common-sense knowledge about prototypical sequences of events and their participants.
Early work focused on structured representations of this knowledge (called {\it scripts}~\citep{schank77}) and manual construction of script knowledge bases. However, these approaches do not 
scale to complex domains~\citep{Mueller98,Gordon01}.
More recently,  automatic induction of script knowledge from text have started to attract attention: these methods exploit either natural texts~\citep{chambers2008unsupervised,chambers2009unsupervised}
or crowdsourced data~\citep{regneri2010learning}, and, consequently, do not require expensive expert annotation.  Given a text corpus, they extract structured representations (i.e. graphs), for example chains~\citep{chambers2008unsupervised} or
 more general directed acyclic graphs~\citep{regneri2010learning}. These graphs are scenario-specific, nodes in them correspond to  events (and associated with sets of potential event mentions) and arcs encode the temporal precedence relation. 
These graphs can then be used to inform NLP applications (e.g., question answering) by providing information whether one event is likely to precede or succeed another.  

In this work we advocate constructing a statistical model
which is capable of  ``answering'' at least some of the questions these graphs can be used to answer, but doing this without explicitly representing the  knowledge as a graph.
In our method,  the distributed representations (i.e. vectors of real numbers) of  event realizations are computed based on distributed representations of predicates and their arguments, and then the event representations are used in a ranker to predict the expected ordering of events.
Both the parameters of the compositional process for computing the event representation and the ranking component of the model are estimated from data. %\footnote{We use text without any kind of semantic annotation on top of them, we call the data unlabeled, in this sense}. 

In order to get an intuition why the embedding approach may be attractive, consider a situation where a prototypical ordering of events \textit{the bus disembarked passengers} and \textit{the bus drove away} needs to be predicted.  An approach based on frequency of predicate pairs~\citep{chambers2008unsupervised}, 
is unlikely to make a right prediction as driving usually precedes disembarking.   Similarly, an approach which treats the whole predicate-argument structure as an atomic unit~\citep{regneri2010learning} will probably fail as well, as such a sparse model is unlikely to be effectively learnable  even from large amounts of data. However, our embedding method would be expected to capture relevant features of the verb frames,  namely, the transitive use for the predicate {\it disembark} and the effect of  the particle {\it away}, and these features will then be used by the ranking component to make the correct prediction.

In previous work on learning inference rules~\citep{Berant11}, it has been shown that enforcing transitivity constraints on the inference rules results in significantly
improved performance. The same is true for the event ordering task, as scripts have largely linear structure, and observing that  $a \prec b$ and $b \prec c$ is likely 
to imply $a \prec c$. 
Interestingly,  in our approach we implicitly learn the model which satisfies transitivity constraints, without the need for any explicit global optimization on a graph.

The approach is evaluated on crowdsourced dataset of \citet{regneri2010learning}  and we demonstrate that using our model results in the 13.5\% absolute improvement in $F1$ on event ordering with respect to their graph induction method (84\% vs. 71\%).

\begin{figure}
\centering
\begin{center}
\includegraphics[width=0.9\columnwidth]{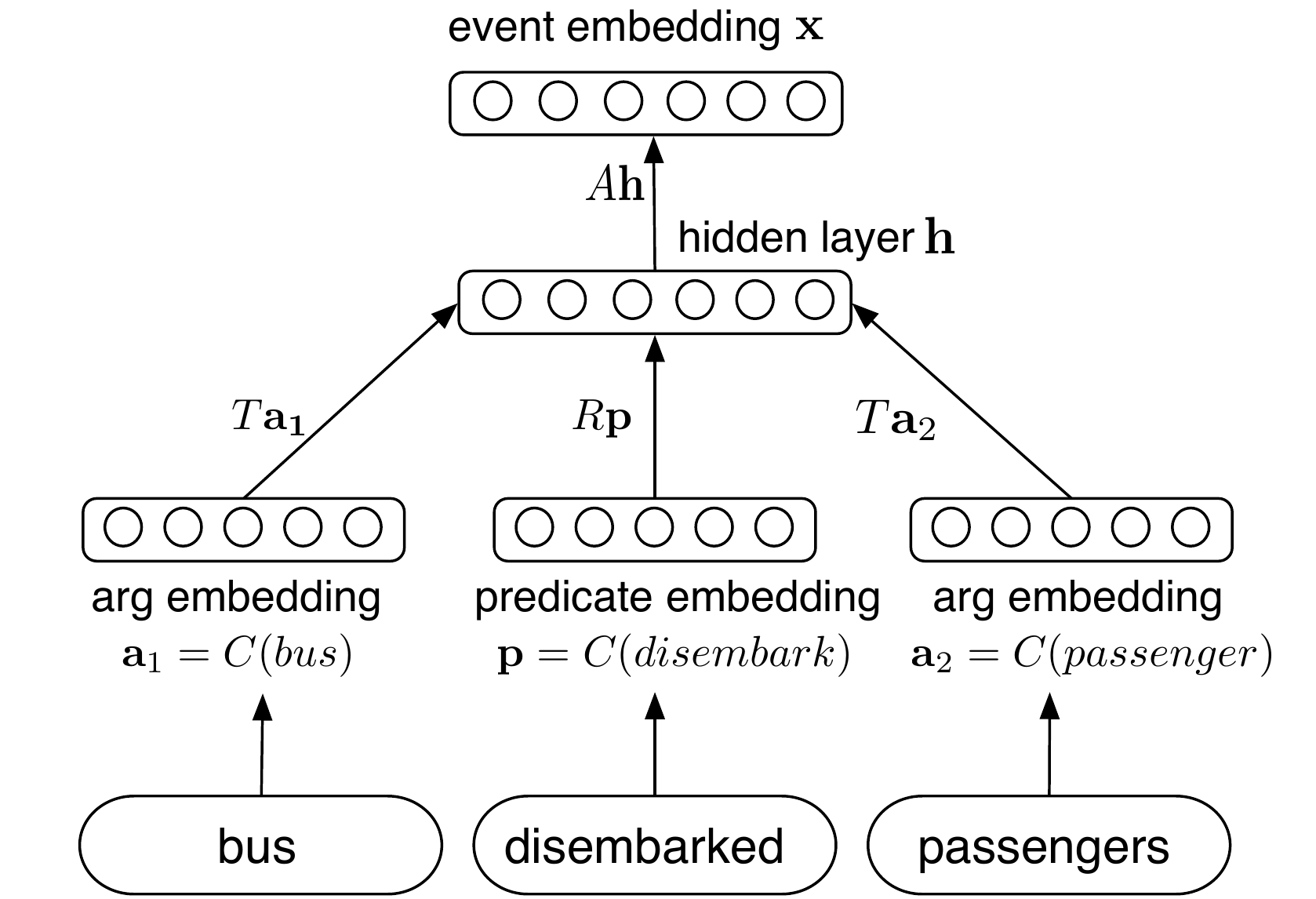} 
\caption{Computation of an event representation ({\it the bus disembarked passengers}).}
\label{model}
\end{center}
\end{figure}

\section{Model}
%\RestyleAlgo{boxed}
\begin{algorithm}[tb]
\caption{Learning Algorithm}
\begin{algorithmic}
%\algrule
\STATE \textbf{Notation}
\STATE $\mathbf{w}:$ ranking weight vector
%\STATE $E:$ sequence of all training event sequences
\STATE ${E_k}:$ $k^{th}$ sequence of events in temporal order
\STATE $t_{k}:$ array of model scores for events in $E_{k}$ 
%\STATE $o_{k}:$ true ordering for event instances in $E_{k}$ 
%\STATE $\mathbf{T}:$ array for training instances for all ESDs
%\STATE $\mathbf{O}:$ array for true order of events in all ESDs
\STATE $\gamma:$ fixed global margin for ranking   
%\end{algorithmic}
%\end{algorithm}

%\begin{algorithm}[H]
%\begin{algorithmic}
%\hrulefill
\algrule
\STATE \textbf{LearnWeights()}
\FOR {$epoch = 1$ to $T$}
\FOR {$k = 1$ to $K$   \hfill[over event sequences]}
\FOR {$i = 1$  to $|E_k|$ \hfill[over events in the seq]}
\STATE Compute embedding $x_{e_{i}}$ for event $e_i$ 
\STATE Calculate score $s_{e_{i}} = \mathbf{w}^{T} x_{e_{i}}$
\ENDFOR
\STATE Collect scores in $t_{k} = [s_{e_{1}},\hdots,s_{e_{i}},\hdots]$
%\STATE insert true order $o_{k}$ in $\mathbf{O} = [o_{1},\hdots,o_{k},\hdots]$
%\STATE sort $t_{k}$  %and insert in $\mathbf{T} = [t_{1},\hdots,t_{k},\hdots]$
%\STATE Update\_Weights($t_{k},o_{k}$)
\STATE $error = \mathbf{RankingError}(t_{k})$ %\COMMENT ranking error for training instance $t_{i}$
\STATE back-propagate $error$ 
\STATE update all embedding parameters and $\mathbf{w}$
\ENDFOR
\ENDFOR
\algrule
\STATE $\mathbf{RankingError}(t_k)$
\STATE $err = 0 $
\FOR {$rank=1,\hdots,l$}
%\STATE $truRank = o_k[idx]$
\FOR {$rankBefore =1,\hdots,rank$}
\IF {$(t_k[rankBefore]-t_k[rank])<\gamma$}
\STATE $err = err + 1$
\ENDIF
\ENDFOR
\FOR {$rankAfter = rank+1,\hdots,l$}
\IF {$(t_k[rank] - t_k[rankAfter]) < \gamma$}
\STATE $err = err + 1$
\ENDIF
\ENDFOR
\ENDFOR
%\RETURN $err$ % doesn't work
\STATE \textbf{return} $err$
%\algrule
\end{algorithmic}
\end{algorithm}

In this section we describe the model we use for computing event representations as well as the ranking component of our model.

\subsection{Event Representation}
\begin{table*}[t] % begin{figure*}[!ht]
%\begin{adjustheight}{2cm}
%\begin{table}
\small{
\begin{center}
\begin{tabular}{|c|ccccc|ccccc|ccccc|}
\hline
 \textbf{Scenario} & \multicolumn{5}{|c|}{\textbf{Precision (\%)}} & \multicolumn{5}{c|}{\textbf{Recall (\%) }} & \multicolumn{5}{c|}{\textbf{F1 (\%)}} \\\hline
 & BL  & EE$_{verb}$ & MSA  & BS&EE    & BL  & EE$_{verb}$ & MSA  & BS&EE   & BL  & EE$_{verb}$ & MSA  & BS&EE\\\hline
  Bus 			& 70.1& 81.9& 80.0&76.0& 85.1  	&71.3 &75.8 &80.0&76.0 & 91.9 	          &70.7 &78.8 &80.0 &76.0& 88.4  \\
  Coffee		& 70.1& 73.7 & 70.0&68.0& 69.5	&72.6 &75.1  &78.0&57.0 &71.0                   &71.3    &74.4&\ 74.0 &62.0 &70.2  \\
  Fastfood 		& 69.9& 81.0 & 53.0&97.0& 90.0 	&65.1&79.1 &81.0&65.0 &87.9    	&67.4 & 80.0 &64.0 &78.0&  88.9  \\
  Ret. Food 		& 74.0& 94.1& 48.0&87.0& 92.4  	&68.6&91.4 &75.0&72.0 &89.7 	         &71.0&  92.8 &58.0 &79.0& 91.0  \\
 Iron			& 73.4&80.1& 78.0&87.0&  86.9	   &67.3&69.8 &72.0&69.0 &80.2      	         &70.2&69.8 &  75.0 &77.0& 83.4 \\
 Microwave	& 72.6&79.2& 47.0&91.0& 82.9	       &63.4&62.8 &83.0 &74.0&  90.3    	&67.7&70.0 &60.0 & 82.0&86.4 \\
 Scr.  Eggs  	 	& 72.7&71.4& 67.0&77.0& 80.7	       &68.0&67.7 &64.0 & 59.0& 76.9		&70.3&69.5 &66.0 & 67.0&78.7 \\
  Shower 		& 62.2&76.2& 48.0& 85.0&80.0    	&62.5& 80.0 &82.0 &84.0&84.3    	&62.3&78.1 &61.0 &85.0&  82.1  \\
 Telephone 	& 67.6&87.8& 83.0&92.0& 87.5	        &62.8& 87.9 &86.0&87.0 &89.0  	         &65.1&87.8 &84.0&89.0 &  88.2  \\
 Vending 		& 66.4&87.3 & 84.0&90.0& 84.2	        &60.6&87.6 &85.0&74.0 &81.9  	&63.3&84.9 &84.0&81.0 & 88.2  \\\hline
\textbf{Average}  	& 69.9&81.3& 65.8&85.0& \textbf{83.9}	   &66.2&77.2 &78.6&71.7 &\textbf{84.3}  	&68.0& 79.1  &70.6&77.6 &\textbf{84.1}   \\\hline
 \end{tabular}
  \caption{Results on the crowdsourced data for the verb-frequency baseline (BL),
  the verb-only embedding model (EE$_{verb}$),  \citet{regneri2010learning} (MSA), \citet{frermannhierarchical}(BS) and the full model (EE).}
   \label{evaluationScriptsFull}  
 \end{center}
}%end of small
%\end{table}
%\begin{adjustheight}
\end{table*}
Learning and exploiting distributed word representations (i.e. vectors of real values, also known as {\it embeddings}) have been shown to be beneficial in many NLP applications~\citep{Bengio01,Turian10,Collobert11}. These representations encode semantic and syntactic properties of a word, and are normally learned in the language modeling setting (i.e. learned to be predictive of local word context), though they can also be specialized by learning in the context of other NLP applications such as PoS tagging or semantic role labeling~\citep{Collobert11}.
More recently, the area of distributional compositional semantics have started to emerge~\citep{Baroni11,Socher12}, they focus on inducing representations of phrases by learning a compositional model. Such a model would compute a representation of a phrase by starting with embeddings of  individual words in the phrase, often this composition process is recursive and  guided by some form of syntactic structure.  

In our work, we use a simple compositional model for representing semantics of a verb frame (i.e. the predicate and its arguments). 
The model is shown in Figure~\ref{model}.
%\footnote{The figure shows a simple example, The model is capable of dealing with more complex sentences as well. For example, consider the sentence "fill water in coffee maker", this contains two phrases as arguments ("water" and "in coffee maker"). For these arguments, we use their "lexical" heads (i.e. "water" and "maker"). The embeddings of these two words and of the predicate ("fill") are then used as the input to the hidden layer. The same procedure is used for predicates with more than two arguments (just more arguments are used as inputs to the hidden layer). In other words, we use a \textit{bag-of-arguments} model.}   
Each word $w_i$ in the vocabulary is mapped  to a real  vector based on the corresponding lemma (the embedding function $C$). 
The hidden layer is  computed by summing linearly transformed predicate and argument embeddings and passing it through the logistic sigmoid function.\footnote{Only syntactic heads of arguments are used in this work. If an argument is \textit{a coffee maker}, we will use only the word {\it maker}. }
We use different transformation matrices for arguments and predicates, $T$ and $R$, respectively. 
The event representation $x$ is then obtained by applying another linear transform (matrix $A$) followed by another application of the sigmoid function.

These event representations are learned in the context of event ranking: the transformation parameters as well as representations of words are forced to be predictive of the temporal order of events. However, one important characteristic of neural network embeddings  is that they can be induced in a multitasking scenario, and consequently can be learned to be predictive of different types of contexts  providing a general framework for inducing different aspects of (semantic) properties of events, as well as exploiting the same representations in different applications.  

\subsection{Learning to Order}
 \label{SubSectLearningToOrder}

The task of learning stereotyped order of events naturally corresponds to the standard ranking setting. 
Here, we assume that we are provided with sequences of events, and our goal is to capture this order. We discuss how we obtain this learning material in the next section. We learn a linear ranker (characterized by a vector $\bs{w}$) which takes an event representation and  returns
a ranking score. Events are then ordered according to the score to yield the model prediction. Note that during the learning stage we estimate not only $\bs{w}$ but also the event representation parameters, i.e. matrices $T$, $R$ and $A$, and the word embedding $C$.  Note that by casting the event ordering task as a global ranking problem we ensure that the model implicitly exploits transitivity of the temporal relation, the property which is crucial for successful learning from finite amount of data, as we argued in the introduction and will confirm in our experiments.

We use an online ranking algorithm based on the Perceptron Rank (PRank,~\citep{Crammer01}), or, more accurately, its large-margin extension. One crucial difference though is that the error is computed not only with respect to $\bs{w}$ but also
propagated back through the structure of the neural network.
The learning procedure is sketched in Algorithm 1. Additionally, we use a Gaussian prior on weights, regularizing both the embedding parameters and the vector $\bs{w}$. 
We initialize word representations using the SENNA embeddings~\cite{Collobert11}.\footnote{When we kept the word representations fixed to the SENNA embeddings and learned only matrices $T$, $R$ and $A$, we obtained similar results (0.3\% difference in the average F1 score).}
%\footnote{\url{http://ronan.collobert.com/senna/}}% We also experimented with the scenario wherein, the word embeddings initialized with SENNA embeddings were kept fixed and not updated during learning. The results for this setting were almost same (F1 score of 84.3\% vs 84.0\% ). In general, keeping them fixed is not a good idea, as learning in the (mostly) language modeling context (as in SENNA) tends to assign similar representations to antonyms/opposites (e.g., open and close). And the opposites tend to appear at different positions in event sequences. However, the fact that the results are similar, may suggest that our dataset is not large enough to learn meaningful refinements. 

\section{Experiments}
\label{SectExp}

We evaluate our approach on crowdsourced data collected for script induction by  \citet{regneri2010learning}, though, in principle, the method is applicable in arguably more general setting of \citet{chambers2008unsupervised}.

\subsection{Data and task}

 \citet{regneri2010learning}  collected short textual descriptions (called {\it event sequence descriptions, ESDs}) of various types of human activities (e.g., going to a restaurant,  ironing clothes) using crowdsourcing (Amazon Mechanical Turk), this dataset was also complemented by descriptions provided in the OMICS corpus~\citep{Gupta04}. The datasets are fairly small, containing 30 ESDs per activity type in average (we will refer to different activities as {\it scenarios}), but the collection  can easily be extended given the low cost of crowdsourcing. The ESDs are written in a bullet-point style and the annotators were asked to follow the temporal order in writing. Consider  an example ESD for the scenario \textit{prepare coffee} :

\noindent
\textit{\{go to coffee maker\} $\rightarrow$ \{fill water in coffee maker\} $\rightarrow$ \{place the filter in holder\} $\rightarrow$ \{place coffee in filter\}  $\rightarrow$ \{place holder in coffee maker\} $\rightarrow$ \{turn on coffee maker\} }

Though individual ESDs may seem simple, the learning task is challenging because of the limited amount of training data, variability in the used vocabulary, optionality of events (e.g.,  going to the coffee machine may not be mentioned in a ESD), different granularity of events and variability in the ordering (e.g.,  coffee may be put in a filter before placing it in a coffee maker). 

Unlike our work, \citet{regneri2010learning}  relies on WordNet to provide extra signal when using the Multiple Sequence Alignment (MSA) algorithm. As in their work, each description was preprocessed to extract a predicate and heads of argument noun phrases to be used in the model.
 
The methods are evaluated on  human annotated scenario-specific tests: the goal is to classify event pairs as appearing in a given stereotypical order or not~\citep{regneri2010learning}.\footnote{The unseen event pairs are not coming from the same ESDs making the task harder: the events may not be in any temporal relation. This is also the reason for using the F1 score rather than the accuracy, both in \citet{regneri2010learning} and in our work.}
 
The model was estimated as explained in Section~\ref{SubSectLearningToOrder}
with the order of events in ESDs treated
as gold standard. 
We used 4 held-out scenarios to choose model parameters, 
no scenario-specific tuning was performed, and the 10 test scripts were not used to perform model selection. 

When testing,  we predicted
that the event pair ($e_1$,$e_2$) is in the stereotypical order ($e_1 \prec e_2$) if the ranking score for $e_1$ exceeded the ranking score for $e_2$

\subsection{Results and discussion}

In our experiments, we compared our event embedding model ({\it EE}) against three baseline systems ({\it BL }, {\it MSA}) and {\it BS}%\footnote{We optimize on accuracy but report results as F1 score. This is so because the binary classification is fairly balanced and there is not much difference between accuracy and F1 scores. Also, we chose the same metric as used in previous work, so that results are comparable.}.
MSA is the system of Regneri et al. (2010). BS is a a hierarchical Bayesian system of \citet{frermannhierarchical}. BL chooses the order of events based on the preferred order of the corresponding verbs in the training set: ($e_1$, $e_2$) is predicted to be in the stereotypical order if the number of times the corresponding verbs $v_1$ and $v_2$ appear in this order in the training ESDs exceeds the number of times they appear in the opposite order (not necessary at adjacent positions); a coin is tossed to break ties (or  if $v_1$ and $v_2$ are the same verb). 

We also compare to the version of our model which uses only verbs (EE$_{verbs}$). Note that EE$_{verbs}$ is conceptually very similar to BL, as it essentially induces an ordering over verbs. However, this ordering  can benefit from the implicit transitivity assumption used in EE$_{verbs}$ (and EE), as we discussed in the introduction. The results are presented in Table~\ref{evaluationScriptsFull}.

The first observation is that the full model improves substantially over the baseline and the previous methods (MSA and BS) (13.5\% and 6.5\% improvement over MSA and BS respectively in F1), this improvement is largely due to an increase in the recall but the precision is not negatively affected.  We also observe a substantial improvement in all metrics from using transitivity, as seen by comparing the results of BL   and EE$_{verb}$ (11.3\% improvement in F1). This simple approach already outperforms the pipelined MSA system. These results seem to support our hypothesis in the introduction that inducing graph representations from scripts may not be an optimal strategy from the practical perspective.

\bibliographystyle{icml2014}
\bibliography{refrences.bib}
\end{document}